%% file: w2v_paper.tex
\documentclass[sigconf,10pt]{acmart}
\acmConference[ICS]{International Conference on Supercomputing}{15-17 June, 2021}{Online}

\usepackage[T1]{fontenc}
\usepackage[utf8]{inputenc}
\usepackage{comment}
\usepackage{algorithm2e} 
	\RestyleAlgo{algoruled}
\usepackage{natbib} 
\usepackage{graphicx} 
\usepackage[font=small,labelfont=bf]{caption} 
\usepackage{listings}
\usepackage{subfigure}
\usepackage{amsmath,amsfonts}
\usepackage{multirow}
\usepackage{xcolor}

\title{FULL-W2V: Fully Exploiting Data Reuse for W2V on GPU-Accelerated Systems}

\copyrightyear{2021}
\acmYear{2021}
\setcopyright{acmlicensed}\acmConference[ICS '21]{2021 International Conference on Supercomputing} {June 14--17, 2021} {Virtual Event, USA}
\acmBooktitle{2021 International Conference on Supercomputing (ICS '21), June 14-118, 2021, Virtual Event, USA}
\acmPrice{15.00}
\acmDOI{10.1145/3447818.3460373}
\acmISBN{978-1-4503-8335-6/21/06}

\begin{document}

\author{Thomas Randall} \affiliation{\institution{Clemson University}\country{USA}} \email{tlranda@clemson.edu}
\author{Tyler Allen} \affiliation{\institution{Clemson University}\country{USA}} \email{tnallen@clemson.edu}
\author{Rong Ge} \affiliation{\institution{Clemson University}\country{USA}} \email{rge@clemson.edu}

\date{\today}

\begin{abstract}
Word2Vec remains one of the highly-impactful innovations in the field of Natural Language Processing (NLP) that represents latent grammatical and syntactical information in human text with dense vectors in a low dimension.
Word2Vec has high computational cost due to the algorithm's inherent sequentiality, intensive memory accesses, and the large vocabularies it represents.
While prior studies have investigated technologies to explore parallelism and improve memory system performance, they struggle to effectively gain throughput on powerful GPUs.

We identify memory data access and latency as the primary bottleneck in prior works on GPUs, which prevents highly optimized kernels from attaining the architecture's peak performance.
We present a novel algorithm, FULL-W2V, which maximally exploits the opportunities for data reuse in the W2V algorithm and leverages GPU architecture and resources to reduce access to low memory levels and improve temporal locality.
FULL-W2V is capable of reducing  accesses to GPU global memory significantly, e.g., by more than 89\%, compared to prior state-of-the-art GPU implementations, resulting in significant performance improvement that scales across successive hardware generations. Our prototype implementation achieves 2.97X speedup when ported from Nvidia Pascal P100 to Volta V100 cards, and outperforms the state-of-the-art by 5.72X on V100 cards with the same embedding quality.
In-depth analysis indicates that the reduction of memory accesses through register and shared memory caching and high-throughput shared memory reduction leads to a significantly improved arithmetic intensity.
FULL-W2V can potentially benefit many applications in NLP and other domains.
\end{abstract}
\maketitle

\input sections/introduction.tex
\input sections/background.tex
\input sections/methodology.tex

\input sections/design.tex

\input sections/results.tex
\input sections/conclusion.tex

\section*{Acknowledgements} 
This work is supported in part by the U.S. National Science
Foundation under Grants CCF-1551511 and CNS-1551262.

\balance
\bibliographystyle{acm}
\bibliography{w2v_paper}

\end{document}

%% file: sections/introduction.tex
\section{Introduction} \label{sec:introduction}

Word embeddings, which represent the meaning of words as numerical vectors, enable computers to understand human language and build efficient complex learning and inferences~\cite{firth}. Word2Vec is a distributed word embedding generator that uses an artificial neural network to learn dense vector representations of words~\cite{google}. The geometry of resulting vectors captures the syntactic and semantic similarities of words, and also exposes complex word and conceptual relationships through vector operations. For example, the words `Rome' and `London' cluster relatively near to one another in the space, and the distance between them is similar in direction and magnitude to the distance between the words `Italy' and `UK'. Word embeddings such as Word2Vec are essential to solving many natural language processing (NLP) problems~\cite{attention, ldt, ws353, simlex999}, including language translation, image captioning, medical software, recommendation systems, various document analysis and have been conceptually expanded to other domains such as graph theory~\cite{node2vec, deepwalk}.

New embeddings are continually needed to capture the latest domain knowledge that can be extracted from ever-growing and ever-evolving corpora and graphs. However useful these word embeddings may be, it is expensive to train new Word2Vec embeddings. First, the Word2Vec algorithm~\cite{google} sequentially trains small moving context windows from the corpus with minimal data parallelism, repeating the process until convergence. Second, the computational complexity scales with both the embedding size and number of unique words to be embedded, the vocabulary, which are ever-increasing for many applications. Third, the state-of-the-art algorithms involve intensive memory accesses and have low arithmetic intensity, limiting hardware scalability. Over time, the demands for high training throughputs continue to increase.
While multiple techniques have been proposed to explore parallelism on GPUs~\cite{blazingtext, parw2v, accSGNS, wombat} and to reduce memory accesses by improving data reuse~\cite{intel, parw2v, pSGNScc}, these works have failed to adequately utilize current-generation GPU hardware and are unlikely to scale to future hardware architectures due to the aforementioned issues. 

\begin{figure}[ht]
\centering
\includegraphics[width=0.8\columnwidth]{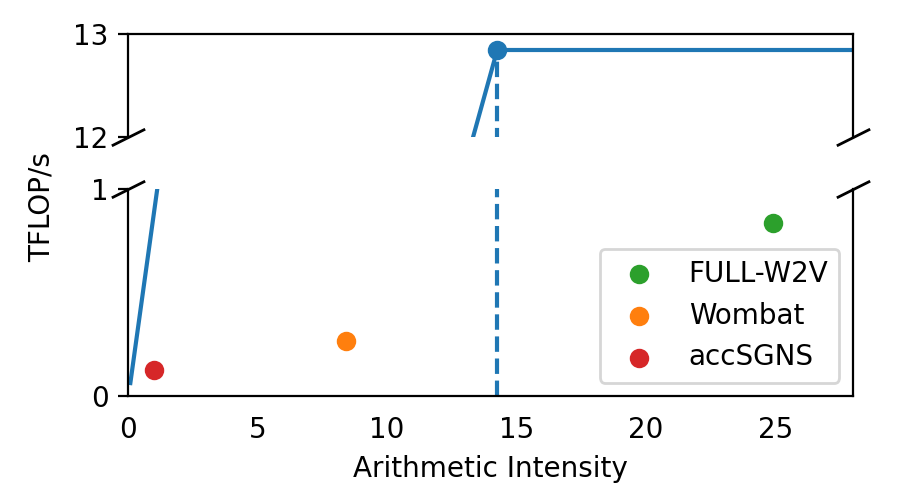}
\caption{Roofline benchmarks for state-of-the-art kernels on a V100 GPU. The solid blue line is the roofline boundary, the 
dotted blue line marks the inflection point between memory-bound (left) and compute-bound (right). Previous work is memory-bound and exhibits poor 
overall throughput despite being data-intensive; our work FULL-W2V presents a significant improvement.}
\label{fig:roofline}
\end{figure}

The state-of-the-art~\cite{blazingtext, wombat, parw2v, accSGNS} GPU implementations of Word2Vec --- Wombat and accSGNS --- 
struggle to effectively utilize the architecture as shown in Figure~\ref{fig:roofline}. While it is well known that 
data-intensive workloads struggle to achieve high arithmetic throughput, GPU implementations of Word2Vec have 
thus far not approached the peak of its potential performance on this architecture.  Our work --- FULL-W2V --- represents
a significant improvements in overall performance and a significant climb in effective arithmetic throughput. 
These results confirm  the challenges in managing latency for Word2Vec on GPUs for data-intensive workloads like Word2Vec as well as 
the necessity of highly-targeted optimizations.

We devise novel technologies to significantly improve Word2Vec performance on GPU architectures. Our key idea is to 
fully exploit  reuse opportunities for different types of words during training, and explicitly cache them in registers 
or shared memory based on their request size and duration.
Word2Vec has a high degree of reuse opportunities that are particularly suited to the storage technologies on GPU if
the algorithm is expressed correctly to take advantage of them. For example, high numbers of available and flexible 
registers, and explicitly allocable shared memory with the same latency as low-level caches.
We take advantage of these technologies to cache reused data fully for the extent of its lifetime, use techniques such
as ring buffers to limit the overhead and management cost of such techniques, and balance heavy storage use to ensure 
scheduling units are still saturated and latency is fully hidden.

In this work, we present FULL-W2V, a fine-grain parallelism, highly scalable Word2Vec GPU algorithm with optimized data 
reuse.  First and foremost, this algorithm maintains the required semantic ordering of context 
windows, maintaining prior guarantees of convergence. Second, it exploits three levels of 
work partition including batches, sentences, and embedding to create high degrees of parallelism on GPUs.
Third, compared to previous work, our algorithm fully exploits  data reuse opportunities, resulting in increased throughput and greater performance scalability, 
nearly eliminating per-thread memory stalls.
Lastly, our algorithm coordinates CPUs and GPUs to seamlessly provision data and launch concurrent kernels to saturate 
GPUs with work.

We have implemented the algorithm and evaluated the prototype on multiple generations of GPUs. Experimental results 
show that our algorithm produces embeddings with the similar quality as existing works. It automatically achieves 
2.966X speedup when moving from Nvidia Pascal 100 and Volta 100 cards. In comparison to the state-of-the-art CPU and 
GPU algorithms, it outperforms state-of-the-art multithreaded CPU implementations by 5.44X, and modern GPU 
implementations accSGNS and Wombat by 5.724X and 8.647X respectively. Deep analysis shows that our algorithm 
increases the arithmetic intensity by 23.90 and 16.46 over accSGNS and Wombat respectively by improving register locality
and utilizing advanced caching techniques to control data reuse.

We make the following contributions in this work:
\begin{itemize}
\item We present FULL-W2V, a fine-grain parallelized, highly scalable Word2Vec GPU algorithm, which overcomes the 
challenges of latency hiding inherent in data intensive Word2Vec training. It achieves 8.647X speedup over the state-of-the-art on Nvidia V100 GPUs.
\item FULL-W2V is the first Word2Vec implementation to exploit \textit{independence of negative samples} to enable opportunities to cache 
and reuse negative samples in registers  for Word2Vec training. It improves arithmetic intensity and instruction 
level parallelism by interleaving memory demand and computation.
\item Realizing that memory access is still a performance bottleneck, FULL-W2V exploits 
\textit{lifetime reuse of context words} to significantly reduce average memory access latency and optimize data sharing, reuse, locality, and coalescing.
\end{itemize}

%% file: sections/background.tex
\section{Background and Related Work} \label{sec:background}

\subsection{Problem Formulation and Algorithm} \label{sec:background:problem}

Word2Vec is a three-layer artificial neural network that learns to represent all words in a vocabulary $V$ as $d$-dimensional vectors $v \in \mathbb{R}^d$ based on their usage in a set of sentences.
These vectors are known as \textit{word embeddings}.
Well-constructed word embeddings can reveal meaningful expressions of syntactic and semantic relationships between words.
For instance, $distance(v_{cat}, v_{dog}) < distance(v_{cat}, v_{hammer})$ indicates that the word ``cat'' is more similar in meaning to ``dog'' than ``hammer,'' and various verb tenses of the same word appear clustered in $\mathbb{R}^d$ to indicate similar syntactic uses.

The Continuous Skip-Gram with Negative Sampling (SGNS) model architecture of Word2Vec has been utilized to great effects in the NLP field for transformers and other higher-level language tasks \cite{ldt}.
The Word2Vec introduced by Mikolov et al.~\cite{google} additionally provides a Continuous Bag-of-Words (CBOW) model architecture for Word2Vec, but Rogers et al.~\cite{ldt} found that the SGNS model architecture generally produces higher quality embeddings for downstream applications of interest, so we focus our attention on the Continuous SGNS model architecture.

\begin{figure}[t]
\begin{centering}
\caption{An example context window (bordered blocks) of size $W=2$ centered on target words in gray. Most words are reused between successive context windows, providing predictable reuse opportunities.}
\includegraphics[width=\columnwidth]{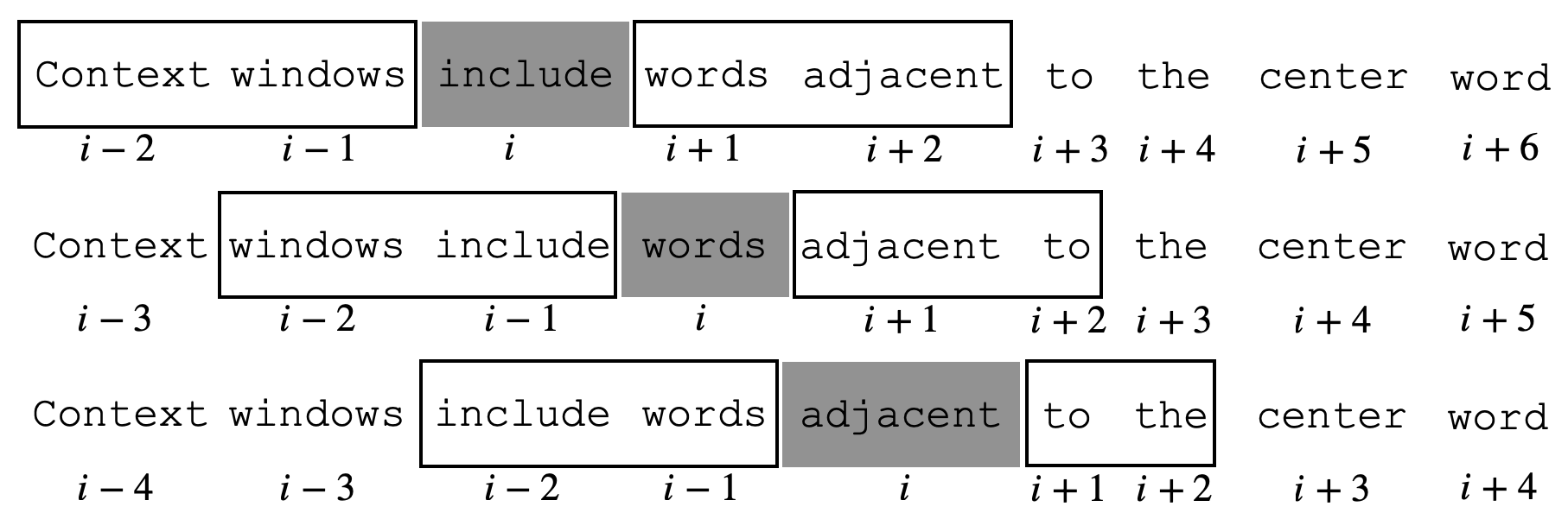}
\label{fig:context}
\end{centering}
\end{figure}

The abstract overview of the Word2Vec algorithm is as follows: the input to Word2Vec is typically a corpus of words that
are organized into ``sentences'', and in each sentence meaningful relationships exist between nearby words; the vocabulary $V$ is formed from
all words in the corpus. In training, the contents of individual sentences are consumed from beginning to end using a sliding
\textit{context window} of $2W$ words where $W$ is the window size, as shown in Figure~\ref{fig:context}.
The center word is the current target word that the model is being trained against. The training algorithm assumes that each word in the context window has increased similarity to the target word as it is  positively related. Such similarity is reflected in the vector representations. The algorithm also assumes that words not present in the context window are \textit{not} related, and their similarity to the target word
decreases.
Rather than decreasing the similarity of all words not included in the context window, SGNS randomly samples  a small number $N$ words using a weighted distribution and makes them more dissimilar to the target word, hence the name ``Continuous Skip-Gram with \textit{Negative Sampling}''. 
These  ``negatives'' greatly reduce the data intensiveness of training as $N << |V|$. Like other neural network models,
SGNS yields a converging solution for word embedding values with a sufficient number of iterations over the data set.

The context window size $W$ and number $N$  of negatives per word are typically defined as hyperparameters 
in  Word2Vec models. Mikolov et al~\cite{mikolov2013efficient, google} established that $W \in [2,10]$ and $N \in [2,20]$ are often
sufficient for most datasets, accelerating training speed without reducing embedding quality, and smaller values of $N$ are more appropriate
for larger datasets.

\input sections/sota.tex

\subsection{Challenges on GPU} \label{sec:background:challenges}

\textbf{Addressing Memory Intensity and Latency.} Like most machine learning algorithms, Word2Vec trains on a large amount of data, has
 inherently low computational intensity, and is generally latency-bound. For CPU implementations, Word2Vec is parallelized coarsely 
along independent sentences. The relatively low computational intensity and throughput of existing GPU implementations have demonstrated that a proper 
decomposition is difficult. To appropriately take advantage of the massive number of cooperative threads and memory hierarchies on the GPU,
 fine-grain parallelism within sentences must also be exploited, so as to effectively minimize high-cost memory accesses, hide latency, 
and maximize the effectiveness of cooperative threads.

\textbf{Managing GPU Resource Tradeoffs.} GPUs have fundamental tradeoffs when using different resources to improve performance. Shared memory and register caches
are both high-speed options for caching data locally for high-locality operations. These two resources are more reliable 
than implicit caching, especially when (1) the required data for many threads may exceed the available footprint in L1 or lower level caches and
 (2) data with different levels of locality are  required for the same computation. However, these resources 
also have limited capacity, and their overallocation can restrict the total
number of resident threads available for execution, leading to  reduced ability to hide latency. Eliminating
expensive memory operations by caching data and hiding latency by using cache to support  thread execution both serve to improve performance, but it is difficult to predict which
is more effective for a given problem. We must tune the usage of these resources to the application's
data locality and balance their usage to maximize performance.

\textbf{Preserving Embedding Quality.}
When exploring new avenues for parallelization and data caching, it is critical that we avoid data dependency violations and minimize race
conditions. All parallel implementations of Word2Vec thus far have had implicit race conditions between sentences containing
the same words, but the impact of this is minimal and does greatly impact the rate at which the algorithm converges 
under the principles described by Hogwild \cite{hogwild}. However, when introducing further parallelization at the sentence
level we risk introducing data dependency violations and compromising the quality of the embeddings produced. Additionally,
changes to allocable memory and explicit caching potentially introduce additional coherency issues that must be managed.
Careful algorithmic analysis and study of data lifetime is required to preserve overall embedding quality.

%% file: sections/sota.tex
\subsection{Related Work} \label{sec:background:sota}

All Word2Vec implementations historically stem from foundational work by Mikolov et al~\cite{mikolov2013efficient,google}, which expresses high level data parallelism between sentences of the corpus for improved performance.
According to Hogwild! SGD~\cite{hogwild}, as long as large models are trained with batches with sufficiently varying contents, parallel gradient descent training can be performed in a lock-free environment without synchronization.
This condition is generally true for Word2Vec with distinct sentences from a given corpus, so data parallelism amongst sentences is commonly exploited using CPU threads or GPU thread blocks.

There have been many implementations of Word2Vec since the seminal works, including implementations for the Tensorflow~\cite{tensorflow} and Gensim~\cite{gensim} machine learning frameworks.
The algorithm has been ported to many architectures, including the cloud-based BlazingText~\cite{blazingtext}, cluster implementation BIDMach~\cite{bidmach}, and FPGA architectures~\cite{fpga}.
We focus the rest of our discussion on published Word2Vec implementations that push the boundaries of the algorithm's throughput on single-node CPU and GPU architectures.

\subsubsection{State-of-the-Art CPU-based Implementations}

\hspace{0pt} \\\indent{}\textbf{pWord2Vec.}
Ji et al~\cite{intel} reduce memory intensity of Word2Vec by ``sharing'' the first $N$ negative samples with all other context words in each window.
For data-intense networks such as Word2Vec, reusing many vectors in each context window's update greatly improves arithmetic intensity, which is further exaggerated by allowing high-performance BLAS libraries to perform the matrix arithmetic.
While the authors were able to show that the semantic changes to the Word2Vec algorithm did not affect embedding quality, the matrix sizes are relatively small and the implementation's performance still fails to approach peak CPU throughput.

\textbf{pSGNScc}
Rengasamy et al~\cite{pSGNScc} utilize advanced batching techniques to combine multiple context windows into larger matrix batches.
The technique allows CPU architectures to achieve much greater throughputs, but computation still takes place entirely on the CPU architecture and is otherwise equivalent in performance to pWord2Vec.

\subsubsection{State-of-the-Art GPU-based Implementations}

\hspace{0pt} \\\indent{}\textbf{accSGNS}
Bae and Yi~\cite{accSGNS} utilize a fine-grain parallel implementation of Mikolov et al's original Word2Vec to bring the algorithm to GPU architectures.
Their parallel hierarchy maps GPU threads directly to embedding layers while thread blocks and grids exploit data parallelism between sentences.
The work's vector parallelism allows for some scalability on newer architectures, but is largely memory latency bound as little is done to affect the data-intensive nature of the Word2Vec algorithm, leading to workload imbalance and poor performance scaling on newer architectures.

\textbf{Wombat}
Simonton~\cite{wombat} focuses on Shared Memory optimizations for Word2Vec, leveraging the architecture's caches to exploit reuse within context windows.
The implementation's parallel formulation uses relatively small thread blocks to operate on fixed word pairings from a context window while grids scale this parallelism across sentences.
The techniques provide state-of-the-art performance on older architectures, but scheduling limitations imposed by the parallel decomposition hold back performance on newer architectures, leaving large room for improvement.

\textbf{PARW2V}
Moon et al~\cite{parw2v} more recently provided CPU and GPU implementations of Word2Vec that induce locality by reordering operations in Word2Vec's training updates and allow for reuse of negative samples beyond a single context window.
The exact degree of negative sample reuse that can be exploited prior to reducing the quality of embeddings was not well understood, and the implementation mandates strict hyperparameter values that limit generalizability.
Furthermore, we were unable to replicate the paper's reported results on our own systems, so this work is not discussed further in this paper.

Our proposed Word2Vec implementation, FULL-W2V, is most related to accSGNS as both works utilize the same parallel hierarchy.
However, we improve upon these techniques by developing a cache for context words that fully reuses them throughout
their lifetime in the sliding window with minimal management, and utilize a local register cache and modified workload breakdown
to gain lifetime reuse of negative samples. All of these methods take advantage of data reuse and problem 
decomposition in ways previously unseen in Word2Vec.

%% file: sections/methodology.tex
\section{Methodology} \label{sec:methodology}

In this section we introduce FULL-W2V, a highly optimized Word2Vec algorithm that is scalable on GPU accelerators.
It overcomes the limited data locality in the state-of-the-art implementations and effectively  exploits GPU architectures in two key ways:
\begin{itemize}
\item it exploits independence of key arithmetic sequences and decouples computations in fine granularity for improved parallelism and reduced data dependency.
\item it fully exploits the temporal locality and data reuse to reduce  access to lower levels of memory and  average memory access time.
\end{itemize}

\subsection{The Independence of Negative Samples }  \label{sec:methodology:negative}

We first introduce the \textit{negative sample independence} property of Word2Vec that allows us to make fine-grain parallelism and highly-effective memory access optimizations.  
When processing each context window in a sentence each context word is paired against each negative, and the sum 
result all pairings is applied as the model update. Because the sum is commutative, 
each pairing may be computed independently in any order.
Acknowledging this independence offers us two opportunities.
First, each negative sample can be independently paired with the context words without synchronization, allowing
\textit{fine-grain parallel} processing among the negative samples.
Second, we can change the order of processing such that all context words are processed for a fixed negative, enabling
\textit{temporal locality} for each negative sample. Recognizing the property 
of negative sample independence, FULL-W2V flexibly manages the order that negatives are processed within a single 
context window and cache them to maximally reduce accesses to low memory levels. 

\textit{Fine-Grain Parallelism and Temporally Distributed Data Dependencies. }
Each individual negative is independently iterated over the context words in a context window,  and the   $N+1$ 
negatives can be fully decoupled from one another.  The decoupling enables two types of opportunities:
(1) fine-grain parallelism 
and (2) reduced simultaneous data dependencies. 
Fine-grain parallelism is crucial to latency hiding and scalable 
performance on GPUs, and provides flexibility for the scheduler to utilize available hardware resources. The 
decoupling reduces the simultaneous data dependency to a single negative sample instead of the whole collection, distributing
the total number of accesses over the lifetime of the computation. Thus it 
eliminates the need for a thread block to simultaneously access and store \textit{all} $N+1$ negatives locally for the duration of
the entire context window. Instead, each thread block only accesses the corresponding negative sample and stores its 
embedding vector directly for its lifetime.

With only one dependent negative, FULL-W2V stores the vector representation in a per-thread register cache. Using registers instead of shared memory has two advantages. First, register access incurs a much 
lower latency than shared memory access and alleviates the demand for latency hiding. Second, a negative sample does 
not have a large number of reuses, which shared memory requires for best cases.  Indiscriminately and aggressively 
using shared memory  reduces the space for thread warps and leads to degraded parallelism, performance, and limits
the quantity that we can use for better-suited optimizations.

\textit{Temporal Locality and Reuse. } FULL-W2V stores each negative sample  in a register  and allows all the required embedding updates in-register before writing it back to memory. Each negative sample is reused by $2W$ times spanning a single  context window.
In this way, we ensure our negative reuse  has minimal impact on the quality of the resultant embeddings. While  prior works~\cite{intel, parw2v, pSGNScc, bidmach} indicate that the particular negative samples  do not need to be independent   across context windows, Moon et al~\cite{parw2v} show that excessive reuse for negative samples has harmful impacts on the final embedding quality. Nevertheless,  the limitations are not well understood by established literature. 
The reuse in a single window  has notable improvement for minimal embedding quality cost~\cite{intel}, and greatly improves the access and storage patterns of negatives for GPU architectures.

One complexity of progressing to the next context window is incremental model updates. 
As the context window slides, context words are reused several times and therefore the corresponding model parameters have data dependencies on prior updates, requiring \textit{strict sequential context window ordering}. In order to adhere to \textit{strict context window ordering} but take advantage of \textit{negative sample independence}, FULL-W2V uses each  thread block to process a full sentence, with individual windows processing all negative samples independently before synchronously sliding the window.
This approach  optimizes the targeted negative reuse without violating any data dependencies or risking over-reusing data.

\subsection{Lifetime Reuse of Context Words} \label{sec:methodology:reuse}

The second optimization enables maximum data reuse for context words  in the algorithmic characteristics of  Word2Vec.
As shown in Figure \ref{fig:context}, we can determine the exact lifetime of context words based on the algorithmic structure of Word2Vec.  
Almost every context word in a given window is also a context word in the subsequent window.
Since successive context windows always shift the  boundary and target word over by one word, every word in the sentence will be a target word once and can appear in up to $2W$ sequential windows as a context word. In other words, a context word's lifetime can be up to $2W+1$ times of reuses. Despite this, existing GPU algorithms fail to realize this degree of reuse or adversely use excessive cache resource by relying on implicit hardware management.  
Separated from existing work, FULL-W2V fully exploits data reuse for the first time, and at runtime explicitly caches and reuses context words with minimum resources.

To reduce expensive high-latency global memory accesses, FULL-W2V  carefully utilizes GPU shared memory  to cache context words for their lifetime.
A naive approach to reusing across multiple windows is to match the size of all words in a context window and allocate space for multiple windows. This approach  would require a prohibitive amount of shared memory, so a more sophisticated and scalable solution is required.
To solve this problem, FULL-W2V builds a circular ring buffer in shared memory to mimic the conceptual sliding window in  Word2Vec.
Each context word vector can be stored in shared memory until the window passes it, allowing the next word to overwrite it.
Using this explicit memory management,  FULL-W2V avoids contention among thread blocks over implicit caches to maximally reuse hot data.
The circular ring buffer also minimizes the amount of shared memory required to store all necessary data for its full lifetime, while  minimizing management overhead.

In order to minimize complexity and maximize  scalability, FULL-W2V alters the implementation of the context width hyperparameter. In traditional Word2Vec, the context width randomly varies between $1$ and $W$
 ~\cite{hyperwords}.
Complying with this treatment of context width would require $2W \times d$ space for a context window, with a full $d$-length vector per word in shared memory, or complex dynamic memory management to handle the context ring buffer.
To simplify \ implementation,  FULL-W2V uses a fixed context width $W_f = \lceil\frac{W}{2}\rceil$, or the average of the original random distribution.
On average, the fixed context width provides the same quality of result while reducing (1) per-context width metadata, (2) the shared memory allocation requirement by half, and (3) the overall implementation complexity of the ring buffer.

With this implementation, FULL-W2V can cache all values in context windows as soon as they appear and accumulate updates in the shared memory until the word is no longer eligible to be a context word.
The overall benefit is a further reduction of global memory accesses by $\frac{2W_f}{2W_f+1}$, approximately 86\% for $W_f=3$, or equivalently a 91\% reduction over $W_f=5$.
In terms of the GPU architecture, this reduces the overall latency and therefore requirement for latency hiding, significantly improving on a key bottleneck.

\subsection{Performance Implications}

\begin{figure*}[htbp]
\centering
\subfigure[FULL-W2V]{\includegraphics[width=0.3\textwidth]{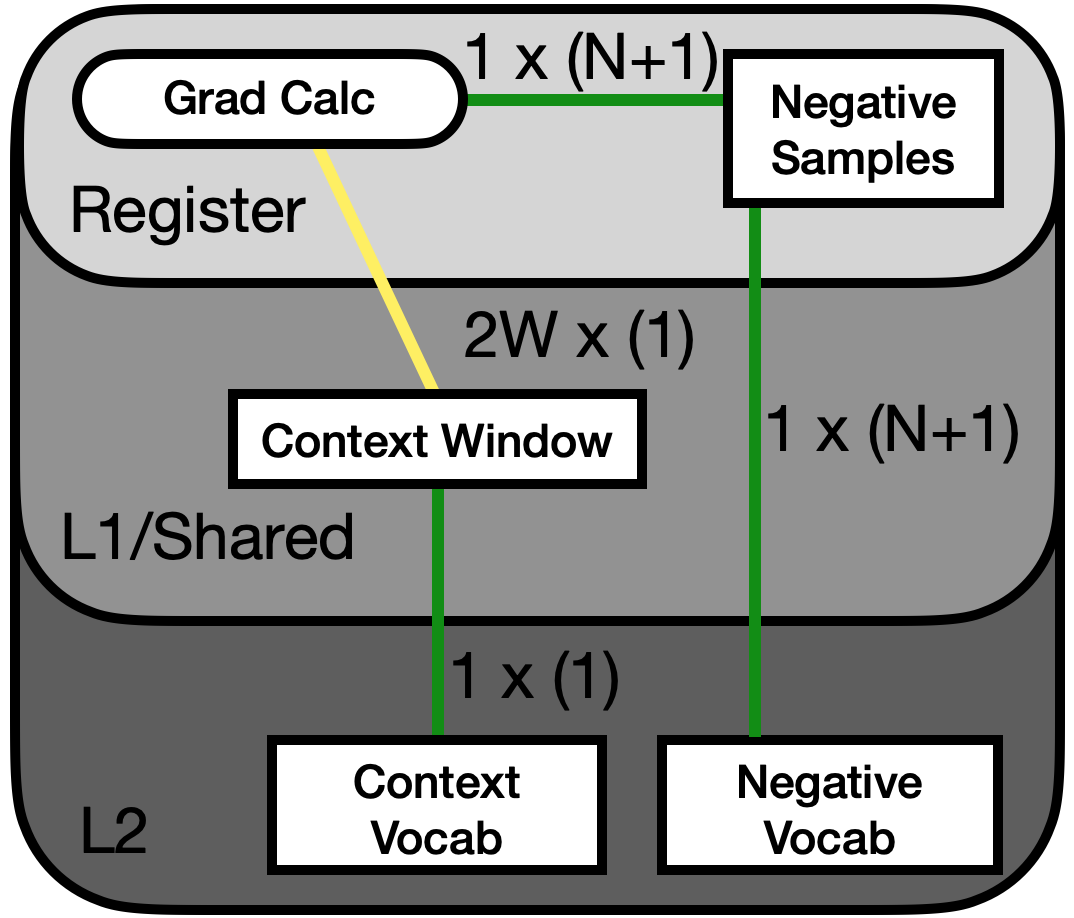} \label{fig:locality:fullw2v}}
\subfigure[Wombat]{\includegraphics[width=0.3\textwidth]{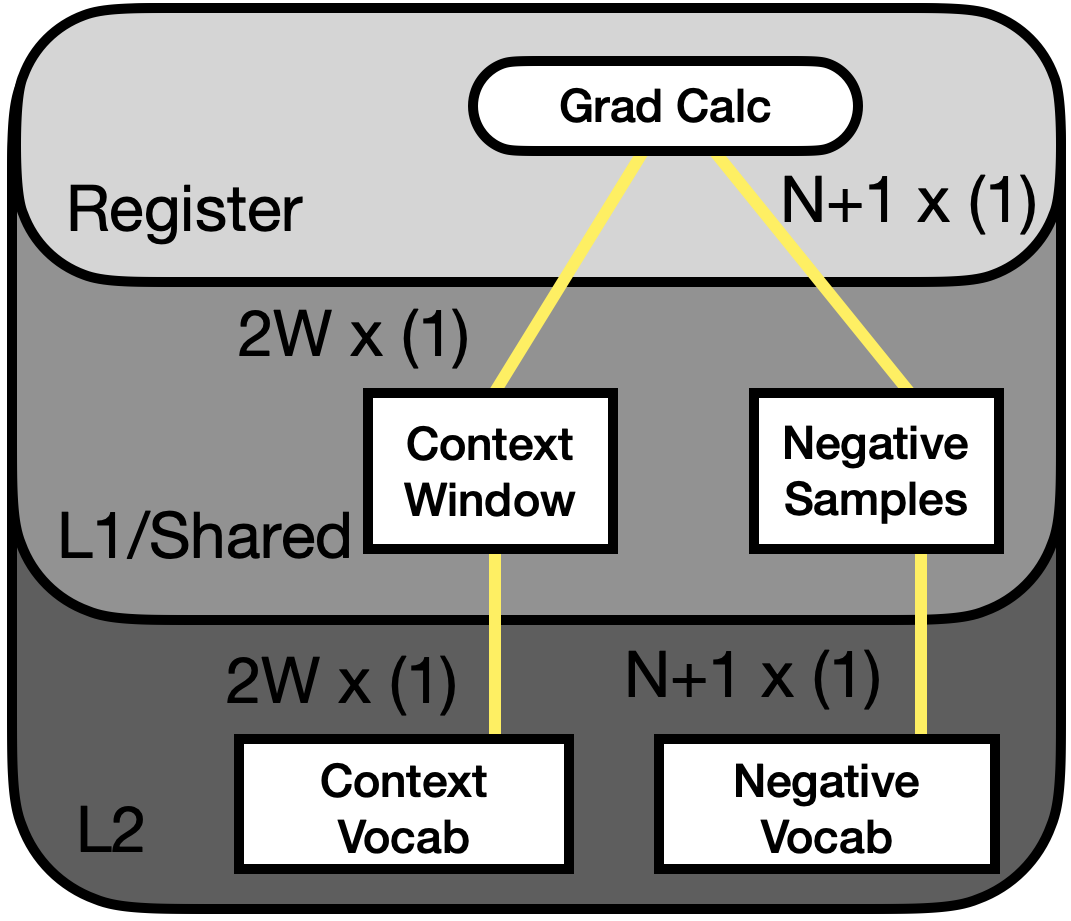} \label{fig:locality:wombat}}
\subfigure[accSGNS]{\includegraphics[width=0.3\textwidth]{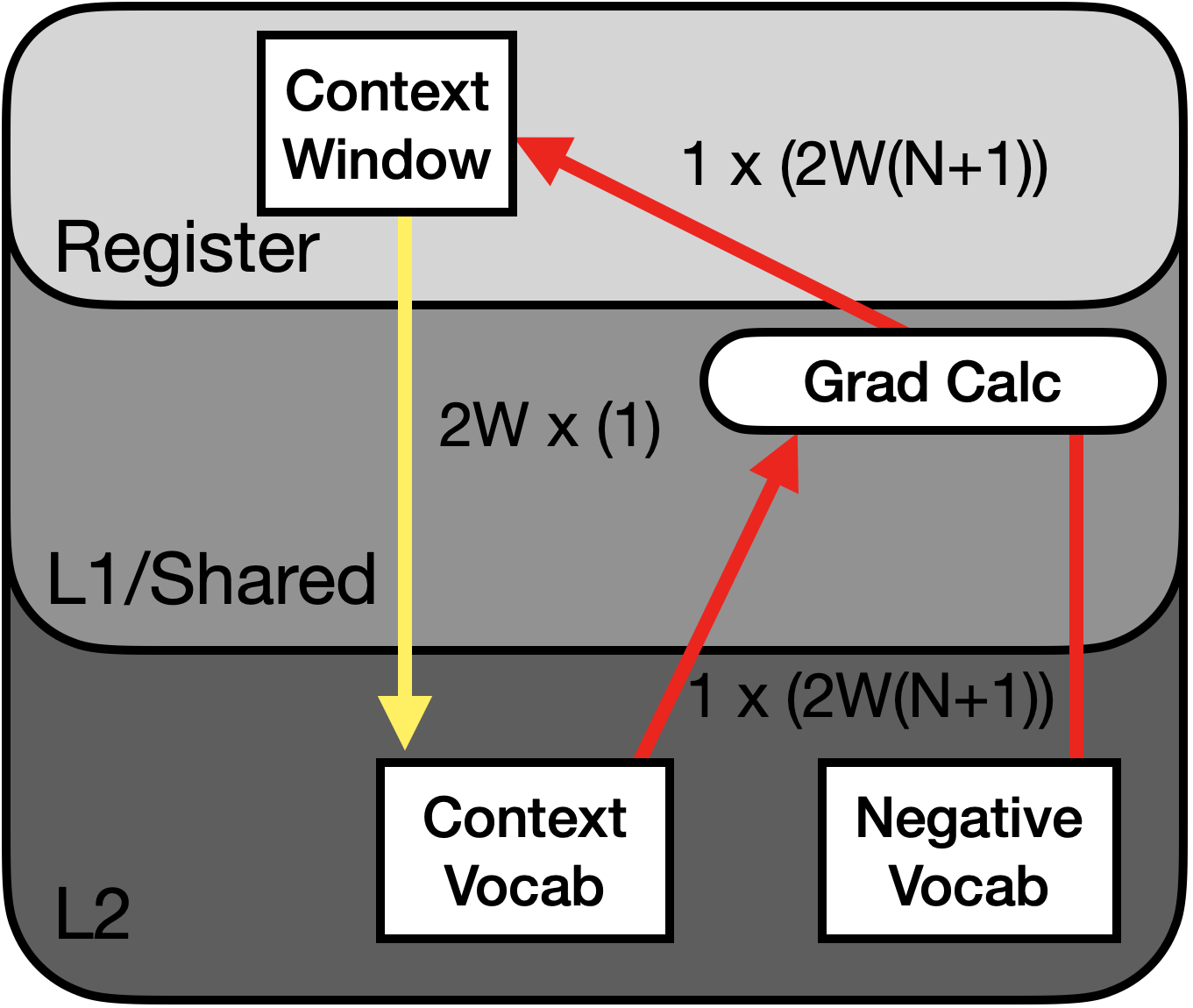} \label{fig:locality:accSGNS}}
\caption{Parallelism and effective data traffic in the memory hierarchy involved in a single context window in the average case. Accesses are shown as $size$ x $(iterations)$, where $2W$ represents the number of context words in a context window and $N+1$ represents $N$ negative samples and the $1$ target word. The colors correspond to how the traffic is related to Wombat (same = yellow, reduced = green, increased = red).}
\label{fig:reuse}
\end{figure*}

By combining fine-grain parallelism and data reuse enabled by the aforementioned technologies, FULL-W2V has significantly 
improved ability to hide memory access latency, and is scalable with GPU architectures. Figure~\ref{fig:reuse} 
summarizes the resulting parallelism and effective data traffic involved in one single context window in the average case, 
where this current window is in the middle of sentence and share context words with multiple precedent and 
subsequent windows. FULL-W2V is distinct from the state-of-the-art Wombat in two key measures.

\begin{itemize}
\item Full-lifetime explicit context and negative caching at the shared memory and register level, respectively.
\item Reduced traffic to each low memory level. FULL-W2V reduces access to L1/shared memory cache by 50\% and access 
to L2 cache and GPU device memory by 42\%, in comparison to Wombat. 
\end{itemize}
 
These differences have several performance implications with GPU architectures and resources. Our method of fine-grain work 
decomposition supports a high degree of parallelism, which is critical for memory intensive workloads such as W2V to 
hide latency and improve instruction-level parallelism. Meanwhile, the thread blocks and warps in FULL-W2V have far
fewer accesses to low level memory and thus memory stalls, improving overall computational efficiency. 
Consequently, FULL-W2V is able to better utilize the computing resource and achieve higher performance. 
Our experimental results  will  provide detailed data to demonstrate the gains. 

FULL-W2V has scalable performance with  generations of GPU architectures. New and more powerful GPUs constantly become 
available. They are typically equipped with more and faster processing units or SMs, more scheduler units, larger caches, 
and higher bandwidth. Given a newer architecture,  FULL-W2V can automatically scale up the degree of parallelism to 
utilize the SMs and provide eligible ready warps to be handled by more scheduler units. Equally importantly,
improved latency hiding, instruction-level  parallelism, and reduced overall memory cost cooperatively improve overall
performance and execution efficiency.

%% file: sections/design.tex
\section{FULL-W2V Design and Implementation} \label{sec:methodology:engineering}

We implement a prototype of FULL-W2V for  test and evaluation of our methodology. The prototype materializes the 
methodology introduced in Section~\ref{sec:methodology} as well as optimize the CPU-GPU coordination.

\subsection{CPU-GPU Coordination} \label{sec:coordination}

\begin{figure}[htbp]
\centering
\includegraphics[width=\columnwidth]{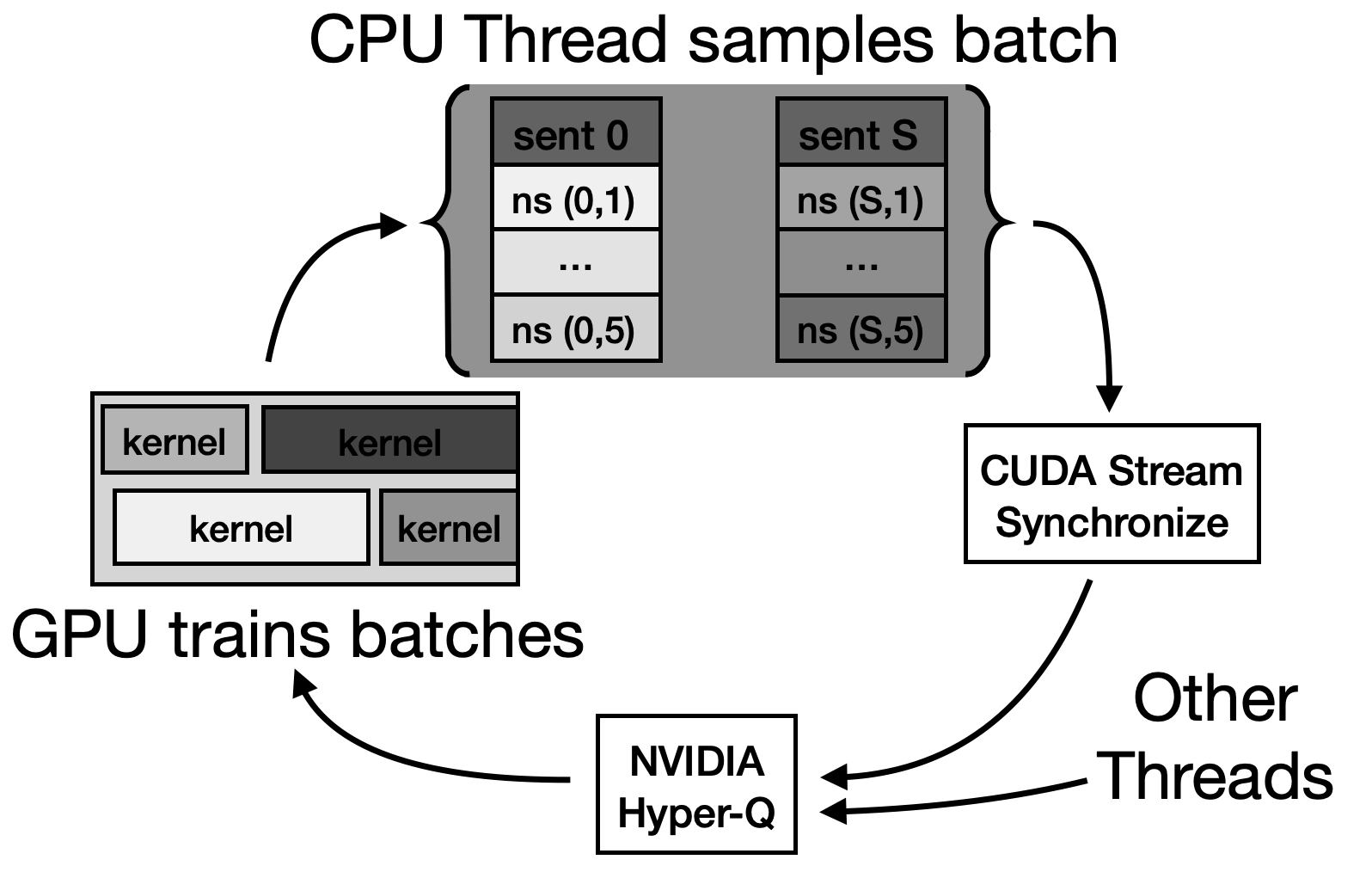}
\caption{The per-stream coordination in FULL-W2V. On each stream, $S = 10,000$ sentences (sent) are sampled from the corpus and $N = 5$ negative samples (ns) are selected for each context window in each sentence.}
\label{fig:cpu-gpu}
\end{figure}

There are two primary goals that require coordination between CPU and GPU devices. The first goal is to ensure that the GPU remains occupied and utilizes its hardware to the greatest possible extent as an accelerator. The second goal is to allocate the workload between devices in such a way that the CPU handles all batch-related precomputation and indirected accesses that would hamper GPU performance if it were instead performed within the kernel.

\textbf{GPU Utilization.} As with other GPU implementations, FULL-W2V partitions the Word2Vec workload into a \textit{batching} component on the CPU, and offloads the batches for \textit{training} on the GPU. In our implementation, batching is precomputation, random sampling, and assembly of data into a format  friendly for GPU, while training is the execution of the Word2Vec algorithm. The heterogeneous coordination is represented in Figure~\ref{fig:cpu-gpu}. Because (1) this is a synchronous process, and (2) batches are relatively small relative to the total computational capability of a GPU, we take advantage of NVIDIA Hyper-Q with CUDA Streams to allow many cooperative CPU threads to batch simultaneously, launching GPU kernels executing in parallel to saturate the GPU.

\textbf{Workload Preprocessing on CPU.} Similar to Wombat~\cite{wombat}, the FULL-W2V batching process includes sentence and negative sample selection. Performing this work on the CPU reduces the number of indirect memory accesses in Word2Vec that need to be performed on the GPU and entirely eliminates GPU copies of several Word2Vec data structures, ultimately improving memory access efficiency on the device. However, unlike Wombat, FULL-W2V does not expand batches into context windows or allow the GPU to reconstruct these windows in the kernel.  It provides indices  as constant memory to the kernel to avoid contention with model memory in the cache hierarchy, making reuse of the data appropriate on the hardware.

Additionally, FULL-W2V adjusts the traditional Word2Vec workload to facilitate more consistent and efficient GPU utilization without impacting model quality. In addition to fixed window width, it  ignores sentence delimiters in training data, thus increasing the average size of sentences and therefore the per-batch workload size. This treatment incurs $<0.5\%$ additional word pairings in common data set but is still worthwhile without adverse performance impact due to better resource utilization.

\begin{table}
\caption{CPU batching speed in millions of words/sec without memory transfers or kernels using the same evaluation conditions as used for overall performance. Batching speed can become a bottleneck for faster implementations of W2V.}
\centering
\footnotesize
\label{table:batching_speed}
\begin{tabular}{|c|c|c|}
\hline \textbf{Implementation} & \textbf{Text8 Batching Speed} & \textbf{1bw Batching Speed} \\\hline
FULL-W2V & 210.340633 & 265.212834 \\\hline
Wombat & 16.957496 & 16.653851 \\\hline
accSGNS & 16.527374 & 15.263448 \\\hline
\end{tabular}
\end{table}

Finally, we note that achieved batching speed is now important for the effective execution of Word2Vec. 
Table \ref{table:batching_speed} represents the rate at which GPU workloads are batched in millions of words-per-second.
Previously, GPU speeds did not approach maximum batching speed. However, with FULL-W2V, we now require the improved 
batching speed demonstrated by our implementation.

\subsection{Parallelism Hierarchy} \label{sec:methodology:parallel}

\begin{figure}[htbp]
\includegraphics[width=1.05\columnwidth]{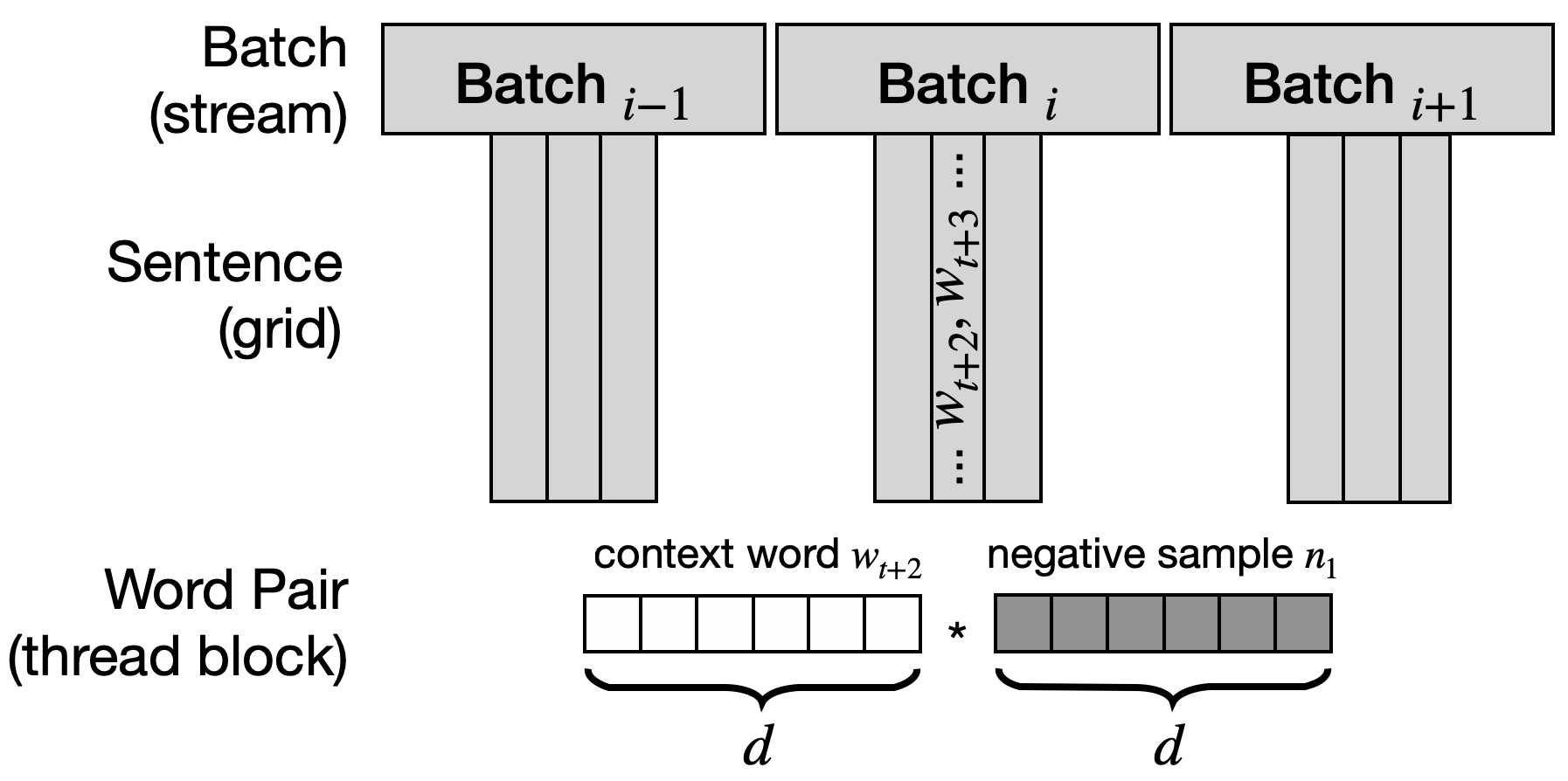}
\caption{The multi-level workload decomposition and parallelism of FULL-W2V. Multiple sentences are batched for each CUDA stream, which launch grids with one thread block per sentence. Each thread block parallelizes embedding layers to operate on pairs of words with many threads.}
\label{fig:hierarchy}
\end{figure}

Maximizing the utilization of GPU architectures demands massive concurrency, particularly for memory intensive 
applications like Word2Vec, to hide data access latency.  FULL-W2V uses a fine-grain, hierarchical parallel approach 
to meet this demand, as shown in Figure~\ref{fig:hierarchy}. The hierarchy consists of three levels of parallel 
Word2Vec training: multi-sentence batch, a sentence and its current context,  and a pair of words or embedding vectors.

\textbf{Batch}: the training corpus is divided into batches and multiple batches are simultaneously trained. This highest level is realized through CPU multithreading and Nvidia Hyper-Q CUDA streams as discussed in Section~\ref{sec:coordination}. FULL-W2V creates one thread per physical CPU core, and each iteratively manages batches and offloads the corresponding training to the GPU via independent streams.

\textbf{Sentence/context window}: There are a number $S$ of sentences in each batch concurrently trained by a  1-d $S$ GPU block grid. We parameterize the number of sentences $S$ per kernel and use $S=10,000$ as an empirical baseline for performance on our systems. Due to the strict sequential context window ordering,  a sentence can only have one current context window, which slides over one word at a step. This context window requires the embedding updates of its context words and the negative samples. In our implementation, we pair one sample with all the context words and calculate the updates, and then iterate over the samples. 

\textbf{Word  pairing}: as each word is represented as a vector, a word pairing involves vector operation, e.g., 
multiplication. The vector computations are parallelized among $d$ threads in the same thread block. This level of 
parallelism enables coalescing and broadcasting of memory accesses, as well as cache availability. Because all threads 
within a block require adjacent vector items, independent warps coalesce their accesses, while potentially making 
the same data available in L1/L2 caches for other warps in collaborating thread blocks.

This hierarchical design can flexibly scale along multiple dimensions to provide strong throughput guarantees under a variety of problem settings and port to new architectures without source code modification. Our prototype implementation is capable of utilizing \textit{word pairing} level scaling to accommodate larger embeddings without modification and automatically gains speedup on architectures with more SMs and warp schedulers.

%% file: sections/results.tex
\section{Experimental Results} \label{sec:results}

In this section we provide our experimental results to quantify FULL-W2V's general performance characteristics and success of our methodology at overcoming the challenges detailed in Section~\ref{sec:background:challenges}.

\subsection{Experimental Platform and Evaluation Method}

We evaluate FULL-W2V on three generations of Nvidia GPUs: V100, Titan XP, and P100 with different processing and memory technologies  in Table~\ref{table:platform}.

\begin{table}[h]
\caption{Evaluation platforms}
\label{table:platform}
\footnotesize
\begin{tabular}{| p{0.24\columnwidth} | p{0.2\columnwidth} | p{0.21\columnwidth} |}
\hline
\textbf{Hardware} & \textbf{GPU Specs} & \textbf{CPU Specs} \\\hline
GPU: V100 \newline Gen-6 Volta \newline \newline CPU: Xeon \newline Gold 6148 \newline Gen-6 Skylake &
80 SMs \newline 14 TFLOP/s \newline 16 GB HBM2 \newline 900 GB/s \newline 4 Warp Sched. &
2 20-core CPUs \newline 2.40 GHz \newline 27.5 MB L3 \\\hline

GPU: Titan XP \newline Gen-5 Pascal \newline \newline CPU: Xeon \newline E5-2670 v3 \newline Gen-4 Haswell &
60 SMs \newline 12.15 TFLOP/s \newline 12 GB GDDR5x \newline 548 GB/s \newline 2 Warp Sched. &
2 12-core CPUs \newline 2.30 GHz \newline 30 MB L3 \\\hline

GPU: P100 \newline Gen-5 Pascal \newline \newline CPU: Xeon \newline E5-2680 v4 \newline Gen-5 Broadwell &
56 SMs \newline 9.3 TFLOP/s \newline 12 GB HBM2 \newline 549 GB/s \newline 2 Warp Sched. &
2 14-core CPUs \newline 2.40 GHz \newline 35MB L3 \\\hline
\end{tabular}
\end{table}

Our analyses compare the following Word2Vec implementations:
\begin{itemize}
\item FULL-Register is a GPU algorithm that implements the techniques described in Sections~\ref{sec:methodology:negative} and~\ref{sec:methodology:engineering}.
\item FULL-W2V is an extension of FULL-Register that additionally implements techniques described in Section~\ref{sec:methodology:reuse}, and represents our full contribution.
\item pWord2Vec~\cite{intel} CPU algorithm is closely related to FULL-W2V and is highly influential on the design of many other Word2Vec works, providing a baseline of expected embedding quality for Word2Vec under Shared Negative Sampling.
\item pSGNScc~\cite{pSGNScc} CPU algorithm has the greatest multicore CPU throughput on our systems with a unique batching mechanism to demonstrate state-of-the-art throughput for Word2Vec on CPU architectures.
\item accSGNS~\cite{accSGNS} GPU algorithm represents a somewhat naive benchmark for CPU-style Word2Vec implemented on GPU hardware.
\item Wombat~\cite{wombat} GPU algorithm provides a state-of-the-art GPU performance for SGNS utilizing shared memory matrix multiplication and in-warp shuffle operations.
\end{itemize}

\textbf{Corpora.}
Following existing literature, we evaluate the quality of generated embeddings using the Text8 corpus~\cite{text8} as well as the One Billion Words corpus~\cite{billionwords}.
Table~\ref{table:corpus} presents summary details regarding each corpus under our experimental conditions.
The Text8 corpus is commonly used for benchmarking evaluations, while the latter includes much more text, allowing it to more reliably predict downstream task performance on a much larger vocabulary~\cite{ldt}.
Therefore we focus on Text8 for throughput analyses and One Billion Words for quality analyses.

\textbf{Evaluation Metrics.}
We evaluate the algorithms with two types of metrics. 

\textit{Training speed and performance.}
We report multiple measures of performance including the training throughput in words per second, and various fine-grain GPU performance data obtained from the $\textit{nsight}$ profiling tool.

\textit{Training quality.}
We utilize spearman's rank coefficient to compare the cosine similarity of word vectors to human similarity judgements established in WS-353~\cite{ws353} and SimLex-999~\cite{simlex999}.
We also use Hyperwords~\cite{hyperwords} to perform analogy reconstruction with cosine addition and multiplication as in the famous Kings-Queens example, and utilize Mikolov's original analogy set~\cite{google} as analogy prompts.

\begin{table}
\caption{Corpus Information. Both corpi only train on words with five or more occurrences in an epoch and are limited to up to 1,000 words per sentence.}
\label{table:corpus}
\footnotesize
\begin{tabular}{| p{0.28\columnwidth} | p{0.18\columnwidth} | p{0.2\columnwidth} | p{0.18\columnwidth}|}
\hline \textbf{Corpus} & \textbf{|Vocabulary|} & \textbf{Words/Epoch} & \textbf{Sentences} \\\hline
\textbf{Text8} & 71,291 & 16,718,845 & 17,006 \\\hline
\textbf{One Billion Words} & 555,514 & 804,269,957 & 30,607,795 \\\hline
\end{tabular}
\end{table}

\textbf{Evaluation Procedure.}
For overall throughput and all embedding quality measures we report the mean and standard deviation of five identical executions to reduce the impact of variance inherent to the Word2Vec algorithm.
All evaluations follow conventional Word2Vec hyperparameters established in Mikolov et al.~\cite{google} with the following noted exceptions.
All experiments use the embedding size of 128, which equalizes GPU performance between all implementations by ensuring each algorithm allocates fully predicated on warps and benefits equally from aligned global memory offsets regardless of the thread block size used by any given kernel--the fundamental performance of FULL-W2V and other GPU algorithms are largely unaffected by this choice.
We allow each implementation to utilize one CPU thread per logical core on the platform.
We allow 20 epochs of training on the Text8 corpus, which was empirically determined to be sufficient for convergence across all implementations; for similar reasons we train the One Billion Words corpus for 5 epochs.

\subsection{Overall Performance} \label{sec:results:perf}

\begin{figure*}[htbp]
\includegraphics[width=1.75\columnwidth]{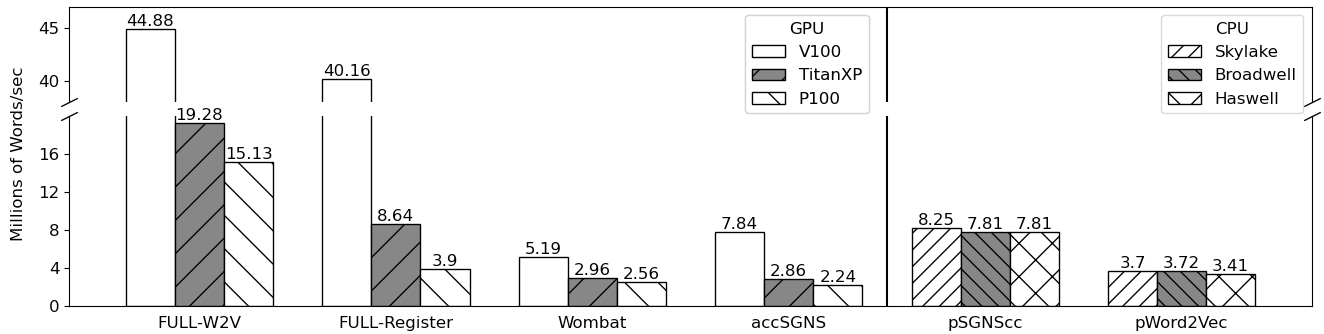}
\caption{Throughput in words/second on Text8 corpus on various architectures. $d=128$, $N=5$, $W=5$.}
\label{fig:architecture_text8_baseline}
\end{figure*}

We first evaluate the overall performance benefits created by our algorithm. 
Figure~\ref{fig:architecture_text8_baseline} shows the training throughput for each algorithm using the Text8 on each experimental architecture.
We make several important observations.

\begin{figure*}[htbp]
\includegraphics[width=1.75\columnwidth]{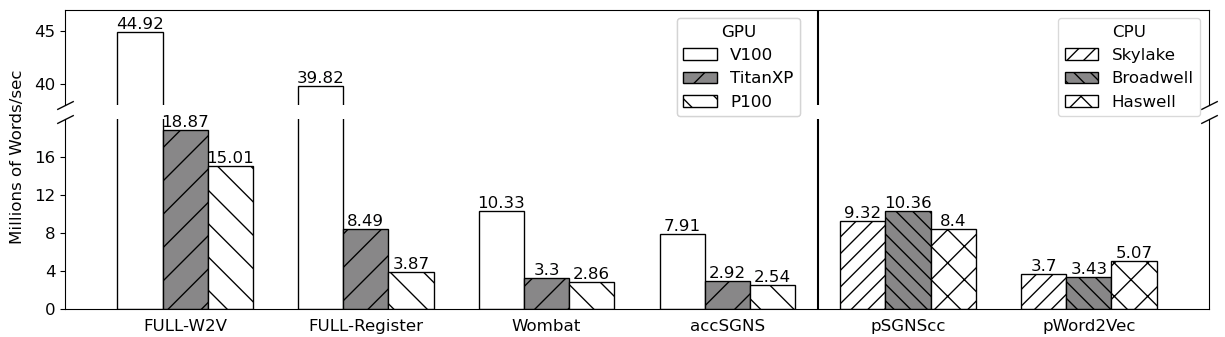}
\caption{Throughput in words/second on One Billion Words corpus on various architectures. $d=128$, $N=5$, $W=5$.}
\label{fig:architecture_1bw_baseline}
\end{figure*}

\begin{itemize}
\item FULL-Register on the XP architecture outperforms all prior works on any architecture and has greater performance scaling cross-architecture than prior works.
FULL-W2V on the P100 nearly doubles FULL-Register's XP performance, but scales its own performance between architectures to a similar degree as prior works.
\item The margin of performance gain for FULL-W2V and FULL-Register over prior works increases with successive hardware generations.
FULL-W2V is 6.754X and 5.910X faster than accSGNS and Wombat respectively on P100 and 5.724X and 8.647X faster than the counterparts on V100.
FULL-Register is 1.741X and 1.523X faster than accSGNS and Wombat respectively on P100 and 5.122X and 7.738X on the V100.
\item Only the FULL-W2V and FULL-Register GPU algorithms are capable of outperforming the peak performance from state-of-the-art CPU algorithms.
AccSGNS on V100 cards achieves comparable performance to the CPU-based algorithms, while Wombat has a lower performance than the pSGNScc algorithm on all three CPUs for the Text8 benchmark and only reaches CPU performance on V100 with the 1bw benchmark.
\end{itemize}

\subsection{Method Evaluation}

\subsubsection{Addressing Memory Intensity and Latency} \label{sec:results:intensity}

\begin{table}
\caption{Memory demand in gigabytes-per-epoch collected via \textit{Nsight} with the Text8 corpus for a fixed number of epochs.}
\centering
\footnotesize
\label{table:memory_demand}
\begin{tabular}{|c|c|c|c|c|c|}
\hline \textbf{Implementation} & \textbf{L1/TEX} & \textbf{L2} & \textbf{DRAM} & \textbf{Sum} \\\hline
FULL-W2V & \textbf{94.760} & \textbf{88.723} & \textbf{41.851} & \textbf{225.334} \\\hline
FULL-Register & 885.065 & 781.576 & 66.555 & 1,733.196 \\\hline
accSGNS & 1,134.448 & 493.614 & 226.578 & 1,854.64 \\\hline
Wombat & 2,303.525 & 1,432.774 & 45.799 & 3,782.098 \\\hline
\end{tabular}
\end{table}

We observe that the register-exploited \textit{independence of negative samples} in FULL-Register results in significant 
reductions in DRAM demand compared to accSGNS, which has a similar access pattern to FULL-Register, eliminating 70.6\% 
of the longest latencies in the memory hierarchy. The lack of register caching and shared negative samples in accSGNS 
leads to much more demand than the lower level caches can satisfy, allowing the hardware cache to achieve some degree of 
success but these accesses are not strictly necessary and proper register allocation leads to demand reduction.
As a data-intensive algorithm, reducing memory demand helps to circumvent the memory latency bottleneck of Word2Vec
on GPU architectures, contributing to the massive performance increases seen in Section~\ref{sec:results:perf} 
between the FULL-Register and FULL-W2V implementations.
The latency bottleneck is more pronounced on older architectures, where fewer SMs with smaller caches and higher 
latency memory technologies expose threads to longer access delays that are otherwise difficult for the 
architecture to hide with other data-intensive work.

We further leverage latency elimination in FULL-W2V, which explicitly manages memory with 
\textit{lifetime reuse of context words} in addition to the negative sample optimizations.
Table~\ref{table:memory_demand} shows that FULL-W2V reduces overall memory demand by 94.0\%, 87.9\%, and 87.0\% 
over Wombat, accSGNS, and FULL-Register respectively.
The extreme reduction in memory demand on other parts of the hierarchy is replaced by reuse in Shared Memory, guaranteeing 
L1 hit latency on each access that is not serviced by the rest of the memory hierarchy.
This is important because the data-intensive access pattern of Word2Vec is both sparse and highly stochastic.
Under these conditions, the GPU's hardware-managed caches cannot be expected to provide proper eviction policies to 
maximize reuse for the algorithm, but \textit{lifetime reuse of context words} guarantees cache hits for the 
Word2Vec algorithm for as long as can be statically known.

\subsubsection{Managing GPU Resource Tradeoffs}

\begin{table*}
\caption{Instructions per Cycle and Thread Stall Breakdown.  Arithmetic stalls include math pipe throttle and MIO. 
Overhead stalls include wait, selection, barriers, dispatch, branch, no instruction, drain, sleep and miscellaneous stalls.
 FULL-W2V shows significant improvements between hardware architectures, and also nearly eliminates memory stalls through effective
manual caching and data reuse. }
\centering
\footnotesize
\label{table:cpi}
\begin{tabular}{|c|c|c|c|c|}
\hline & \multicolumn{2}{c|}{\textbf{XP}} & \multicolumn{2}{c|}{\textbf{V100}} \\\hline
							& FULL-Register	& FULL-W2V & FULL-Register	& FULL-W2V \\\hline
\textbf{IPC} 					 & 1.19  	& 2.78	& 2.38 	& 3.22 \\\hline
\textbf{Long Scoreboard}			& 38.66		& 1.25	& 11.00		& 0.97 \\\hline 
\textbf{Short Scoreboard}			& 4.49		& 3.43	& 4.19		& 2.95 \\\hline 
\textbf{Arithmetic}				& 0.16		& 0.18	& 1.14		& 0.66 \\\hline 
\textbf{Overhead}				& 13.05		& 10.48	& 7.93		& 6.35 \\\hline 
\end{tabular}
\end{table*}

\begin{table*}
\caption{Average Issue Eligibility per Warp Scheduler per Cycle. Maximum active warps on both architectures is 16. FULL-W2V
is always near-peak occupancy and has near-ideal eligible warps, indicating good latency hiding as well as scheduler
saturation. }
\centering
\footnotesize
\label{table:eligible}
\begin{tabular}{|c|c|c|c|c|c|c|c|c|}
\hline & \multicolumn{4}{c|}{\textbf{XP}} & \multicolumn{4}{c|}{\textbf{V100}} \\\hline
				& Wombat		& accSGNS	& FULL-Register	& FULL-W2V & Wombat	& accSGNS	& FULL-Register	& FULL-W2V \\\hline
\textbf{Max Warps}	& 11.03		& 12			& 16			& 13		& 11.03		& 12			& 16			& 9 \\\hline
\textbf{Active Warps} & 4.59		& 11.08		& 15.86		& 9.59	& 4.66		& 9.41		& 14.92		& 8.99 \\\hline
\textbf{Eligible Warps} & 0.16		& 1.33		& 0.42		& 0.99	& 0.18		& 1.09		& 1.86		& 1.90 \\\hline
\end{tabular}
\end{table*}

We analyze our effectiveness in managing resource tradeoffs by examining scheduler and thread-level statistics, starting 
with device processor and scheduler saturation.
Table~\ref{table:eligible} shows that FULL-W2V is within $99\%$ of its theoretical occupancy, indicating both
inter-SM and intra-SM saturation of threads. Additionally, we can see that active warps are near-peak levels with 
appropriate (near $1$) eligible warps-per-scheduler. This indicates that many warps are progressing on some operation, while a 
sufficient number of warps are available for scheduling. High activity and balanced eligible warps is a good indication that
our latency reduction operations eliminated a sufficient amount of latency to justify a lowered overall occupancy without harming 
our ability to hide the remaining latency and still improve overall performance. This is also an indication that we can continue
to scale to future architectures, as we are not approaching any hardware limitations and scaling to new SMs is simply 
tied to batching additional sentences. 

We validate that our method reduces overall time spent on latency, we look at lower-level per-thread metrics, including
overall IPC and its constituent breakdown. Table \ref{table:cpi} shows that, despite improved performance, FULL-Register
still spends a great number of cycles stalled on latency costs, particularly long scoreboard memory operations. 
On both architectures, the introduction of \textit{lifetime reuse of context words} nearly eliminates the cost of 
long-access memory, indicating that we nearly eliminate this cost. In turn, IPC is drastically increased, shifting 
much of the remaining time to compulsory overhead operations, including synchronization. This single-thread 
improvement is highly validating of our overall throughput gains.

\subsubsection{Preserving Embedding Quality.}
\begin{table}
\caption{Mean embedding quality of five repeated trials using One Billion Words. Higher values are better.}
\label{table:architecture_quality_baseline}
\footnotesize
\begin{tabular}{|c|c|c|c|c|c|c|}
\hline \textbf{SW} & \textbf{WS-353} & \textbf{SimLex-999} & \textbf{COS-ADD} & \textbf{COS-MUL} \\
\hline \textbf{pWord2Vec} & 0.6070 & 0.3499 & 29.895\% & 29.166\% \\
\hline \textbf{Wombat} & 0.5952 & 0.3596 & 29.661\% & 28.988\% \\
\hline \textbf{FULL-W2V} & 0.5923 & 0.3582 & 29.775\% & 29.386\% \\
\hline
\end{tabular}
\end{table}

We evaluate the embedding quality of FULL-W2V and compare it against Wombat and pWord2Vec as presented in Table~\ref{table:architecture_quality_baseline}.
These counterparts use the same batching semantics and negative sample reuse policies as FULL-W2V and thus create a fair comparison.
FULL-W2V is statistically equivalent to the results generated by both Wombat and pWord2Vec for every measure of the training quality.

This positive result confirms that our algorithmic adjustments described in Section~\ref{sec:methodology:reuse}, including fixed context window sizes, 
are valid. As demonstrated in~\cite{strangegeometry}, larger window sizes are connected to divergence in learning quality 
between high and low-frequency words, but variance in window sizes does not appear to be critical to generating quality embeddings.

%% file: sections/conclusion.tex
\section{Conclusion and Future Work} \label{sec:conclusion}

FULL-W2V advances the state-of-the-art single-GPU performance  across multiple hardware generations.
We find that each negative sample in a collection can be independently updated over context words without affecting embedding quality, however the sequential accumulation of context word updates throughout sliding windows remains necessary for convergence.
Based on these findings, we improve the efficiency of fine-grain parallelism with highly effective memory access optimizations --- cache negatives in registers and context words in shared memory --- to fully exploit their reuse. We show that the combination of fine-grain parallelism, novel memory demand reductions, and data reuse optimizations can generate synergistic performance gains and benefits on GPU hardware.

There are several directions for future work.
There is a lack of understanding of the exact limitations of negative sample reuse without adversely affecting embedding quality.  FULL-W2V and future algorithms can benefit from  reuse of negatives over more than one context window. Related work shows that  altering sentence batching and negative sample selection  increases limits of guaranteed locality for additional performance benefits. FULL-W2V is positioned to explore such benefits. 
Finally, FULL-W2V can be extended to support multiple GPUs on the same node to further accelerate training and support large networks and corpus.

%% file: w2v_paper.bbl
\begin{thebibliography}{10}

\bibitem{accSGNS}
{\scshape Bae, S., and Yi, Y.}
\newblock Acceleration of word2vec using gpus.
\newblock In {\em Proceedings of the 23rd International Conference on Neural
  Information Processing\/} (10 2016), vol.~9948, pp.~269--279.

\bibitem{bidmach}
{\scshape {Canny}, J., {Zhao}, H., {Jaros}, B., {Chen}, Y., and {Mao}, J.}
\newblock Machine learning at the limit.
\newblock In {\em 2015 IEEE International Conference on Big Data (Big Data)\/}
  (2015), pp.~233--242.

\bibitem{billionwords}
{\scshape Chelba, C., Mikolov, T., Schuster, M., Ge, Q., Brants, T., and Koehn,
  P.}
\newblock One billion word benchmark for measuring progress in statistical
  language modeling.
\newblock {\em CoRR abs/1312.3005\/} (2013).

\bibitem{ws353}
{\scshape Finkelstein, L., Gabrilovich, E., Matias, Y., Rivlin, E., Solan, Z.,
  Wolfman, G., and Ruppin, E.}
\newblock Placing search in context: The concept revisited, 2001.

\bibitem{firth}
{\scshape Firth, J.~G.}
\newblock A synopsis of linguistic theory 1930-1955" in studies in linguistic
  analysis.

\bibitem{node2vec}
{\scshape Grover, A., and Leskovec, J.}
\newblock node2vec: Scalable feature learning for networks.
\newblock {\em Proceedings of the 22nd ACM SIGKDD International Conference on
  Knowledge Discovery and Data Mining\/} (2016).

\bibitem{blazingtext}
{\scshape Gupta, S., and Khare, V.}
\newblock Blazingtext: Scaling and accelerating word2vec using multiple gpus.
\newblock In {\em Proceedings of the Machine Learning on HPC Environments\/}
  (New York, NY, USA, 2017), MLHPC’17, Association for Computing Machinery.

\bibitem{simlex999}
{\scshape Hill, F., Reichart, R., and Korhonen, A.}
\newblock Simlex-999: Evaluating semantic models with (genuine) similarity
  estimation.
\newblock {\em Computational Linguistics 41}, 4 (2015), 665--695.

\bibitem{intel}
{\scshape {Ji}, S., {Satish}, N., {Li}, S., and {Dubey}, P.~K.}
\newblock Parallelizing word2vec in shared and distributed memory.
\newblock {\em IEEE Transactions on Parallel and Distributed Systems 30}, 9
  (Sep. 2019), 2090--2100.

\bibitem{hyperwords}
{\scshape Levy, O., Goldberg, Y., and Dagan, I.}
\newblock Improving distributional similarity with lessons learned from word
  embeddings.
\newblock {\em Transactions of the Association for Computational Linguistics
  3\/} (2015), 211--225.

\bibitem{text8}
{\scshape Mahoney, M.}
\newblock Large text compression benchmark.
\newblock Unpublished paper, 2011.

\bibitem{mikolov2013efficient}
{\scshape Mikolov, T., Chen, K., Corrado, G., and Dean, J.}
\newblock Efficient estimation of word representations in vector space, 2013.

\bibitem{google}
{\scshape Mikolov, T., Sutskever, I., Chen, K., Corrado, G.~S., and Dean, J.}
\newblock Distributed representations of words and phrases and their
  compositionality.
\newblock In {\em Advances in Neural Information Processing Systems 26},
  C.~J.~C. Burges, L.~Bottou, M.~Welling, Z.~Ghahramani, and K.~Q. Weinberger,
  Eds. Curran Associates, Inc., 2013, pp.~3111--3119.

\bibitem{strangegeometry}
{\scshape Mimno, D., and Thompson, L.}
\newblock The strange geometry of skip-gram with negative sampling.
\newblock In {\em EMNLP\/} (2017).

\bibitem{parw2v}
{\scshape {Moon}, G.~E., {Newman-Griffis}, D., {Kim}, J., {Sukumaran-Rajam},
  A., {Fosler-Lussier}, E., and {Sadayappan}, P.}
\newblock Parallel data-local training for optimizing word2vec embeddings for
  word and graph embeddings.
\newblock In {\em 2019 IEEE/ACM Workshop on Machine Learning in High
  Performance Computing Environments (MLHPC)\/} (2019), pp.~44--55.

\bibitem{hogwild}
{\scshape Niu, F., Recht, B., Re, C., and Wright, S.~J.}
\newblock Hogwild!: A lock-free approach to parallelizing stochastic gradient
  descent, 2011.

\bibitem{fpga}
{\scshape Ono, T., Shoji, T., Waidyasooriya, H.~M., Hariyama, M., Aoki, Y.,
  Kondoh, Y., and Nakagawa, Y.}
\newblock Fpga-based acceleration of word2vec using opencl.
\newblock {\em 2019 IEEE International Symposium on Circuits and Systems
  (ISCAS)\/} (2019), 1--5.

\bibitem{deepwalk}
{\scshape Perozzi, B., Al-Rfou, R., and Skiena, S.}
\newblock Deepwalk: online learning of social representations.
\newblock In {\em KDD '14\/} (2014).

\bibitem{pSGNScc}
{\scshape Rengasamy, V., Fu, T.-Y., Lee, W.-C., and Madduri, K.}
\newblock Optimizing word2vec performance on multicore systems.
\newblock In {\em Proceedings of the Seventh Workshop on Irregular
  Applications: Architectures and Algorithms\/} (New York, NY, USA, 2017),
  IA3'17, Association for Computing Machinery.

\bibitem{ldt}
{\scshape Rogers, A., Hosur~Ananthakrishna, S., and Rumshisky, A.}
\newblock What's in your embedding, and how it predicts task performance.
\newblock In {\em Proceedings of the 27th International Conference on
  Computational Linguistics\/} (2018), Association for Computational
  Linguistics, pp.~2690--2703.

\bibitem{wombat}
{\scshape {Simonton}, T.~M., and {Alaghband}, G.}
\newblock Efficient and accurate word2vec implementations in gpu and
  shared-memory multicore architectures.
\newblock In {\em 2017 IEEE High Performance Extreme Computing Conference
  (HPEC)\/} (Sep. 2017), pp.~1--7.

\bibitem{tensorflow}
{\scshape Tensorflow}.
\newblock Tensorflow.
\newblock \url{https://www.tensorflow.org/tutorials/text/word2vec}.

\bibitem{attention}
{\scshape Vaswani, A., Shazeer, N., Parmar, N., Uszkoreit, J., Jones, L.,
  Gomez, A.~N., Kaiser, L., and Polosukhin, I.}
\newblock Attention is all you need.
\newblock {\em ArXiv abs/1706.03762\/} (2017).

\bibitem{gensim}
{\scshape Řehůřek, R.}
\newblock Gensim.
\newblock \url{https://radimrehurek.com/gensim/models/word2vec.html}.

\end{thebibliography}
